\documentclass[journal,transmag]{IEEEtran}
\ifCLASSINFOpdf
\else
\fi

\usepackage{amsmath}
\usepackage{hyperref}
\usepackage{url}
\usepackage{amsmath}
\usepackage{graphicx}
\usepackage{lipsum}

\begin{document}
%
\title{A Generative Adversarial Model for Right Ventricle Segmentation}


\author{\IEEEauthorblockN{Nicol\'o Savioli \IEEEauthorrefmark{1} Miguel Silva Vieira\IEEEauthorrefmark{1},
Pablo Lamata\IEEEauthorrefmark{1},
Giovanni Montana\IEEEauthorrefmark{1} \IEEEauthorrefmark{2}}
\IEEEauthorblockA{\IEEEauthorrefmark{1}King's College London, Division of Imaging Sciences and Biomedical Engineering, UK}
\IEEEauthorblockA{ \IEEEauthorrefmark{2} WMG, University of Warwick, UK}\\
\texttt{\IEEEauthorrefmark{1}\{nicolo.l.savioli,miguel.silvavieira,pablo.lamata\}@kcl.ac.uk} \\
\IEEEauthorrefmark{2}\texttt{g.montana@warwick.ac.uk} }

%
\IEEEtitleabstractindextext{%
\begin{abstract}
\end{abstract}
}
\maketitle
\IEEEdisplaynontitleabstractindextext
%
\IEEEpeerreviewmaketitle

\begin{abstract}
The clinical management of several cardiovascular conditions, such as pulmonary hypertension, require the assessment of the right ventricular (RV) function. This work addresses the fully automatic and robust access to one of the key RV biomarkers, its ejection fraction, from the gold standard imaging modality, MRI. The problem becomes the accurate segmentation of the RV blood pool from cine MRI sequences. This work proposes a solution based on Fully Convolutional Neural Networks (FCNN), where our first contribution is the optimal combination of three concepts (the convolution Gated Recurrent Units (GRU), the Generative Adversarial Networks (GAN), and the L1 loss function) that achieves an improvement of 0.05 and 3.49 mm in Dice Index and Hausdorff Distance respectively with respect to the baseline FCNN. This improvement is then doubled by our second contribution, the ROI-GAN, that sets two GANs to cooperate working at two fields of view of the image, its full resolution and the region of interest (ROI). Our rationale here is to better guide the FCNN learning by combining global (full resolution) and local Region Of Interest (ROI) features. The study is conducted in a large in-house dataset of $\sim$ 23.000 segmented MRI slices, and its generality is verified in a publicly available dataset.
\end{abstract}

\section{Introduction}

Cardiovascular diseases (CV) remain the leading cause of death worldwide \cite{KreatsoulasC}, accounting for 17.3 million total deaths worldwide. In the management of these conditions, cardiac magnetic resonance (CMR) is considered the gold standard for the assessment of key biomarkers such as the volume or ejection fraction (EF) of the ventricular chambers of the heart \cite{Stacey}. The fully automatic and robust access to this important diagnostic and prognostic information is nevertheless missing in the clinical armamentarium. 

The Left Ventricle (LV) has traditionally focused the clinical interest for the characterization of the disease progression, but in recent years a strong shift of attention to the Right Ventricle (RV) has led to important findings for the management of conditions such as pulmonary hypertension, coronary heart disease, dysplasia and cardiomyopathies \cite{RobertNaeije,Davlouros}. 

Compared to the LV, the RV is a much more challenging anatomical structure to be characterized, mainly because of a much larger morphological variability and the much thinner myocardial walls \cite{Petitjean}. 
The RV biomarker of volume, EF or cardiac output is conventionally accessed through the acquisition of a stack of short-axis (SA) slices of the heart. The problem of interest then becomes the automatic RV segmentation in CMR SA slices, where the goals are the removal of the (intro- and inter-) observer variability and the immediate access to this information right after acquisition. 

The attention on RV segmentation was initiated with a-priori probabilistic atlases \cite{LorenzoValdes, Mitchell}, using both shape and appearance information. The strength and weakness of this approach lay on the suitability of the cohort used to build the atlas: this solution will render a low performance in new anatomical configurations not accounted in the training dataset. In an attempt to alleviate this limitation, manifold learning techniques have been applied to better capture the variability of shape models, for example using Markov Random Field (MRF)\cite{Moolan}. Image gradient algorithms\cite{Grosgeorge}, region-merging\cite{Maier} and graph-cut\cite{Mahapatra} based methods have been shown to be more compelling. The implementation of these ideas led to popular concepts such as active contours able to reach reasonable performance, but with a dependence on the actual choice of weighting factors and the optimal initialization point.  

In the last few years, deep-learning (DL) methods are being developed for extracting automatic spatial features. Particularly,  Fully Convolution Neural Networks (FCNN) can be considered the state of the art or the automatic segmentation of the RV \cite{PhiVT, GongningLuo, LiemanSifry}. 

In this work we want to further extend the capability of FCNNs, exploring two main ideas: first, modeling and exploiting the spatial redundancy between adjacent SA slices; and second, guiding the FCNN to the useful RV features without the need of an automatically pre-localisation of the RV region of interest (ROI). The approach to exploit spatial redundancy is the incorporation of a recurrent unit in the middle of up-sampling and down-sampling path of the FCNN, an R-FCNN, a strategy that has been shown to improve the LV segmentation, especially at the apex \cite{RudraP}. 

Guiding an FCNN to the correct image features is a much more complex goal. The explicit ROI extraction is an approach followed by RV segmentation \cite{Avendi2016} and many medical applications \cite{Shreyas, WangZehan, Seokwon} to facilitate the segmentation task. Mask-RCNN \cite{Kaiming} is an example where the segmentation obtained from an FCNN is in close combination with an ROI-pooling mechanism able to locally identify the bounding box of each object. Our rationale is that there are still useful image features outside the ROI that can guide the FCNN, and that approaches that jointly learn detection and segmentation are desirable, avoiding the only focus on the ROI features. Some works explore this idea, where ROI pre-localisation becomes an additional sequential task in an end-to-end training chain \cite{Marvin, BlitzNet}, but without an explicit use for guiding the segmentation. While a dual FCNN within local and global downsampling pathways at two different MRI resolution was used for atrial segmentation problem \cite{ZXiong}.  However, in this work, the local path only helps to scan every single patch of the image in order to classify it as negative or positive. In truth, this method differs from the principle of FCNN (i.e uses downsampling filters to scan the whole pixel image) and approaches more to prior old segmentation techniques, where the adding the global path works as a multiscale integration of global contexts. 

The strategy to guide the FCNN to features within the ROI, without losing the features outside it, is inspired in the concept of the Coupled Generative Adversarial Networks (CoGAN) \cite{Ming1,Ming2}, where a pair of corresponding images in different domains can be mapped in the same representation within a shared parameters strategy between two Generative Adversarial Networks (GAN). We adopt this concept by taking two versions of the same image, one at full resolution and another at the exact ROI around the segmentation mask, and we call it the ROI-GAN. Note that the second image will only be needed at the training stage.

In this work, we thus explore the use of three existing concepts, a recurrent unit (R-FCNN) to exploit the spatial redundancy of a stack of SA slices, the concept of adversarial training (FCNN-GAN) to better guide the selection of features, and the use of the L1 loss \cite{DPathak, PhillipIsola}. And we propose the ROI-GAN as a solution to maximize the performance of FCNNs for the task of RV segmentation.

\section{Material and methods} 

In this section, we present the datasets used in this work, and we review the concepts of the FCNN, the R-FCNN, the L1 loss, and the GAN training strategy using either FCNN or an R-FCNN. Finally, our ROI-GAN architecture is explained, with 3 possible variants that will be analyzed.

\subsection{Datasets} 

Two datasets are used, a large in-house Twins-UK dataset to exhaustively develop and test the DL solution, and a small public RV MICCAI dataset  \cite{Petitjean} to test the generality of findings.  

Twins-UK is a nation-wide voluntary registry that includes \textgreater 12,000 twins \cite{MoayyeriA}. The study was approved by the local institutional research ethics committee (South East London Research Ethics Committee, EC04/015), and informed consent was obtained from all participants prior to scanning. In particular, 68 consecutive female patients (mean age $62 \pm 9$ years) were recruited from the Twins-UK cohort. 

All scans were performed on a 1.5-T clinical scanner (Achieva, Philips Healthcare, Best, The Netherlands). All measurements were performed in the supine position using a 5-channel cardiac surface coil. ECG-gated steady-state free-precession (SSFP) end-expiratory breath-hold 2D CINE were acquired. 12 to 14 equidistant and contiguous slices from the atrioventricular (AV) ring to the apex, completely covering both ventricles (slice thickness 8 mm; no gap mm; field of view was $360 \times 480$ mm and matrix size $156 \times 144$), were acquired. The imaging was performed a temporal resolution of 25-35 milliseconds at a heart rate of 60-80 beats per minute. 

The RV endocardial borders were delineated in all planes and in all cardiac phases by Dr. Miguel Silva Vieira (6 years of CMR experience, SCMR and Euro CMR level 3 certification). In detail, the ventricular blood pool was segmented using a semi-automated threshold-based contouring technique, 
which enables to capture details of the endocardial configuration (e.g trabeculations). Of note, papillary muscles were excluded from the volumetric analysis (equivalent to blood pool techniques).

The Twins-UK dataset of ~23.000 slices with segmentation ground truth was randomly divided into training, validation and testing sets of sizes $70\%$, $15\%$ and $15\%$, respectively. 

On the other hand, the public RV MICCAI dataset was used to refine weights (16 subjects with two-time points segmented, ~250 slices) and to evaluate performance on the Test2Set blind cohort used for benchmarking (another set of 16 subjects, a similar number of slices). This public dataset is composed of subjects with a variety of disease conditions, with an average age of $55.5 \pm 17.5$ and where $70\%$ of them were male.

\subsection{The FCNN/R-FCNN}

An FCNN is the core of the DL solution, taking a stack of SA slices as input and returning the segmentation mask. An FCNN extract features from the image in a first decoding path, and these are gradually restored to the original image size, through a decoding block used to infer the final binary mask. One of the key features of an FCNN are the skip paths connections between convolution and deconvolution layers for fusing mid-height level features together. Each connection concatenates feature maps from the encoding to the decoding blocks. 

\begin{figure*}[ht]
\centering
 \includegraphics[width=7in]{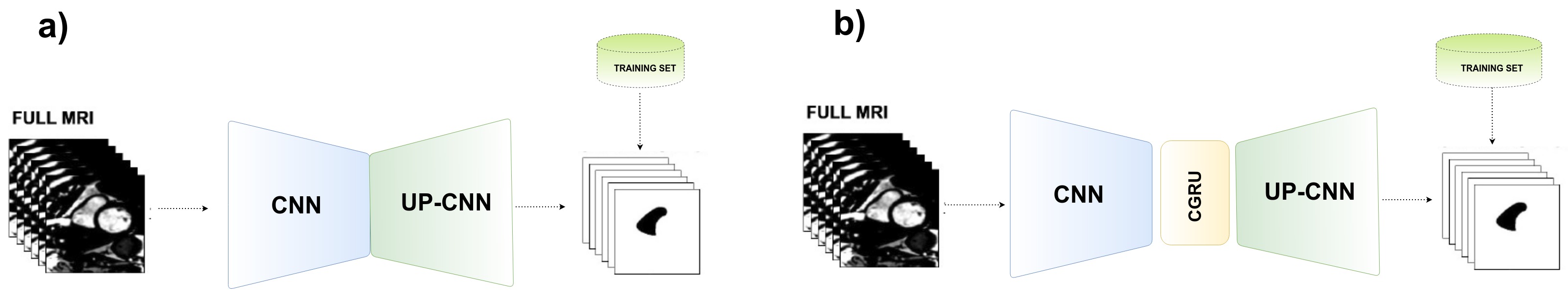}
   \caption{FCNN and RFCNN architectures. The FCNN (panel a) is a combination of a decoder and encoder paths. The decoder path (blue trapezoid) consists of six convolution layers (stride of 2) following by ReLU and Batch Normalization (BN). The up-convolution path (green trapezoid) consists of deconvolution layers in combination with LeakyReLU (set to $0.2$) and Batch Normalization. The R-FCNN (panel b) is similar to FCNN but a convolution-GRU (yellow rectangle) is used in between the decoder and encoder in order to model and exploit the spatial MRI redundancy.}
  \label{fig:model_1}
\end{figure*}

The FCNN (Fig. \ref{fig:model_1} panel (a)) considers each slice of the SA input independently. The extension in an R-FCNN (Fig. \ref{fig:model_1} panel (b)) is to take the full SA stack as an input, as a sequence of slices from base to apex where the current slice depends on the previous observed ones \cite{RudraP}. 

The R-FCNN has the same encoding and decoding structure of an FCNN (i.e six encoding and decoding blocks followed by ReLU and Batch Normalization (BN) operators) but, in the middle of both, a Convolution Gated Recurrent Unit (C-GRU) \cite{Junyoung, MennatullahSiam} is used. The C-GRU unit presents two specific gates designed to control the information inside: a rest gate $r_{s}$ and an update gate $z_{s}$ defined as follow: 

\begin{align*}
  r_{s} & = \sigma (W_{hr} * h_{s-1} + W_{xr}* x_{s} + b_{r}) \\
  z_{s} & = \sigma (W_{hz} * h_{s-1} + W_{xz}* x_{s} + b_{z})
\end{align*}

Here, $\sigma (\cdot)$ is the sigmoid function and the $h_{s-1}$ represents the hidden activation learned at the previous SA slice $s-1$.
The $W_{hr}$ and $W_{hz}$ are the weight matrices of dimension $D \times 8 \times 8$, with D numbers of features maps in the down-sample layer, and $b$ a bias vector.

In this notation, $*$ defines the convolution operation. 
The reset gate switch (on or off) the signal coming in input to $\hat{h_{s}}$; where  $\hat{h_{s}}$ is called the candidate activation, defined as:

$$
\hat{h}_t = \tanh(W_{h}*(r_{s} \odot h_{s-1}) + W_{x} *x_{s} + b) 
$$

The $\odot$ denotes the dot product and $W_{x}$. Then the final activation is:

$$
h_{s} = (1-z_{s}) \odot  h_{s-1} + z_{s} \odot \hat{h_{s}} .
$$

\subsection{The L1 loss}

The training of a network is guided by the metric used to define the error, and the L1 loss \cite{DPathak,PhillipIsola} has shown to be a good addition to the total loss. The L1 distance used is given as:

\begin{equation}
 L_{L1} =\frac{\beta}{n}  \mid x_{i}-y_{i} \mid
\end{equation}

This metric measures the mean absolute value of element-wise difference among the network output $x^{i}$ and ground truth $y^{i}$, where, $\beta$ is a regularization constant parameter (set to $5e^{-6}$) for controlling the quantity of L1 loss used.

\subsection{The GAN}

\begin{figure*}[!ht]
\centering
 \includegraphics[width=7in]{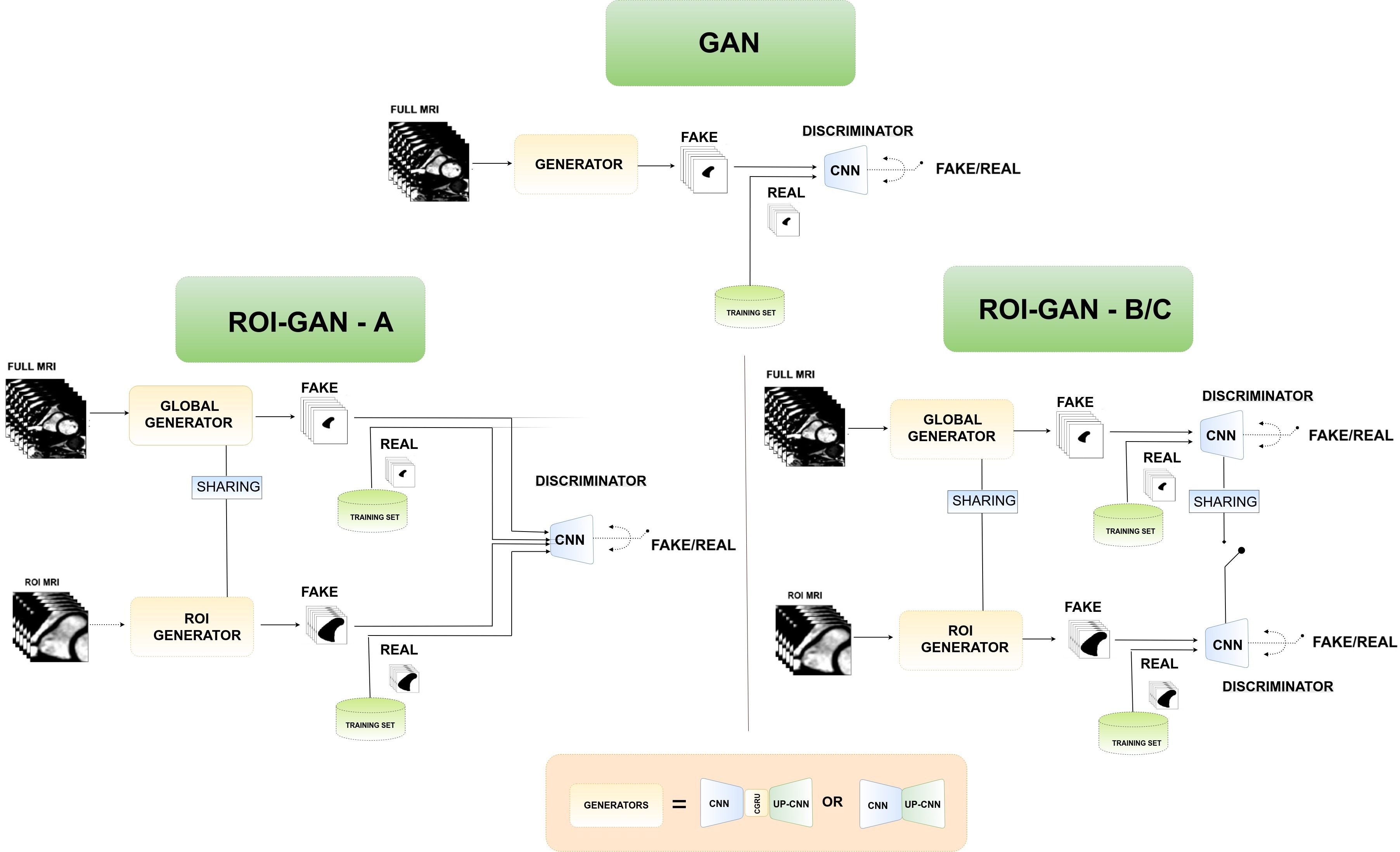}
   \caption{The three ROI-GAN architectures studied in this work. TOP: basic GAN architecture. BOTTOM LEFT: the ROI-GAN-A, where masks at two different sizes are feeding the same discriminator CNN. BOTTOM RIGHT: the ROI-GAN-B/C architectures, where two different CNN are used as discriminators, one for each image size, either in coordination (i.e. sharing parameters) in B configuration, or independently in C configuration.} 
 \label{fig:model_2}
\end{figure*}

Two neural net architectures, the generative and the discriminative, compete in a GAN to perform a task. We adopt this concept to our problem so that the FCNN or R-FCNN become generators of binary masks (note that their input here is not a distribution of random numbers, but the distribution of MRI images), and we add a new CNN to act as discriminator that will try to identify if the binary mask is ``fake''  (i.e. output of the generator) or ``real" (i.e. the ground truth mask). See the top of Fig. \ref{fig:model_2} for an illustration of a GAN architecture.
 
The discriminator CNN network is thus trained to distinguish how much the ground truth deviates from that produced by the FCNN generator.
This information generated by the discriminator is back-propagated towards the generator, which then uses this knowledge for producing indistinguishably masks from the corresponding ground truth. Besides, in order to avoid deterministic generators, Gaussian Noise (GN) is added by dropping the first three up-sampling layers of the generator.

The adversarial process is summarized by the maximization of the following loss:

\begin{equation}
  \begin{aligned}
      min_{G} \thinspace \thinspace max_{D} L_{GAN}(D,G) = \\ 
      =  E_{x \sim {MRI_{real}(x)}}[log(D(x))]  + \\ 
      +  E_{x \sim {MRI_{fake}}(x)}[log(1-D(G(x)))]
  \end{aligned}
\end{equation}

where $D$ represent the discriminator, $G$ the discriminator and $x$ is the set of binary 
masks sampling form $MRI_{real}(x_{k})$ and $MRI_{fake}(x)$. 

This adversarial loss is also combined within the minimum squared error $MSE$ loss function between the generator output and the ground truth:

\begin{equation}
  L_{MSE}(G) =\frac{1}{n} \sum_{i=1}^{n} (x_{i}-y_{i})^{2}
\end{equation}

where $n$ is the set of training spatial sequences cases (i.e. slices), $x$ is the generated binary mask and $y$ is the ground truth segmentation. The two loss functions are combined through a $\lambda$ regularization parameter (set to $5e^{-3}$) able to control the amount of GAN loss taken into account:

\begin{equation}
\begin{aligned}
L_{TOTAL}(G,D) = (     L_{MSE}(G) + \lambda  L_{GAN}(D,G)   )
\end{aligned}
\end{equation}

\subsection{The ROI-GAN}

The ROI-GAN model (Fig. \ref{fig:model_2}), takes inspiration from the concept of CoGAN \cite{Ming1, Ming2}, which is adapted to working with the same image but at two different fields of view: one a the global level (i.e. original full resolution MRI image), and another at the region of interest (ROI) local level. We will thus refer to the global (working with the full resolution) or local (working with the cropped image) generators and discriminators in each of the two collaborative GANs. 

Note that the cropped images needed for the second set of images, the ROIs, are simply the bounding boxes containing the ground truth segmentation, and are only needed for the training phase of the architecture. 

The idea is that the local GAN will inform and better guide the global GAN. And this is articulated through a mechanism of parameter sharing between the generators and discriminators of the two GANs in an attempt to intensify the attention on the correct subset of mid-level features. 

The training process is sequential: first, the cropped MRI images are segmented (i.e forward pass on the local generator) with a corresponding backward propagation of the loss by comparison to the segmentation ground truth. Second, the updated parameters of the local generator are passed to the global generator by using the weights-sharing connections that are enabled on the first three up-sampling layers. Then, the global generator repeats the forward and backward process. The third step is the training of the discriminator and the backpropagation of the total network gradient from the discriminator to both generators.

Three different strategies (identified by the surname A, B or C) for the discriminator third step are designed and compared in this work. ROI-GAN-A (Fig. \ref{fig:model_2} bottom left) uses one single discriminator fed by the images at both fields of view (full and ROI), motivated by the idea of maximizing the interplay between the two generators by sharing the same discriminator. The alternative is to use two discriminators (Fig. \ref{fig:model_2} right), one per generator, in the ROI-GAN-B. The third option is an intermediate solution, where the two discriminators are allowed to share weights between them, as set in the ROI-GAN-C.

\section{Results} 

\begin{figure*}[ht]
\centering
 \includegraphics[width=7in]{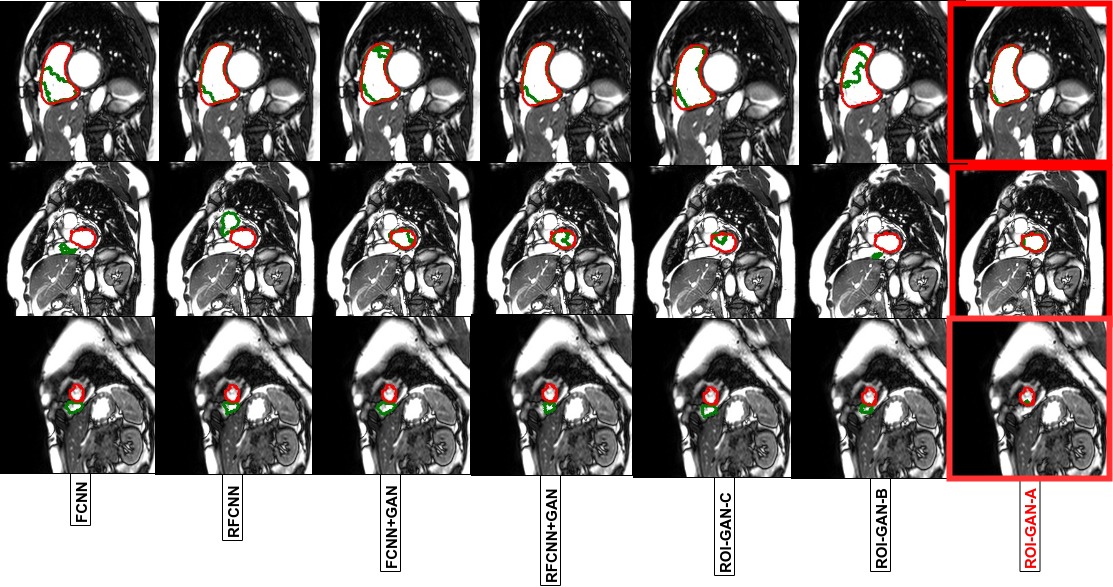}
   \caption{Illustrative segmentation results on our in-house Twins-UK dataset, comparing neural networks predictions (green line) to the ground truth (red line). The ROI-GAN-A (last column) shows a good match both in an easy case (first row, a slice from the top of the RV) and in a difficult case (third row from the apical low region or the RV).}
  \label{fig:twinsuk}
\end{figure*}

\begin{figure*}[ht]
\centering
 \includegraphics[width=7in]{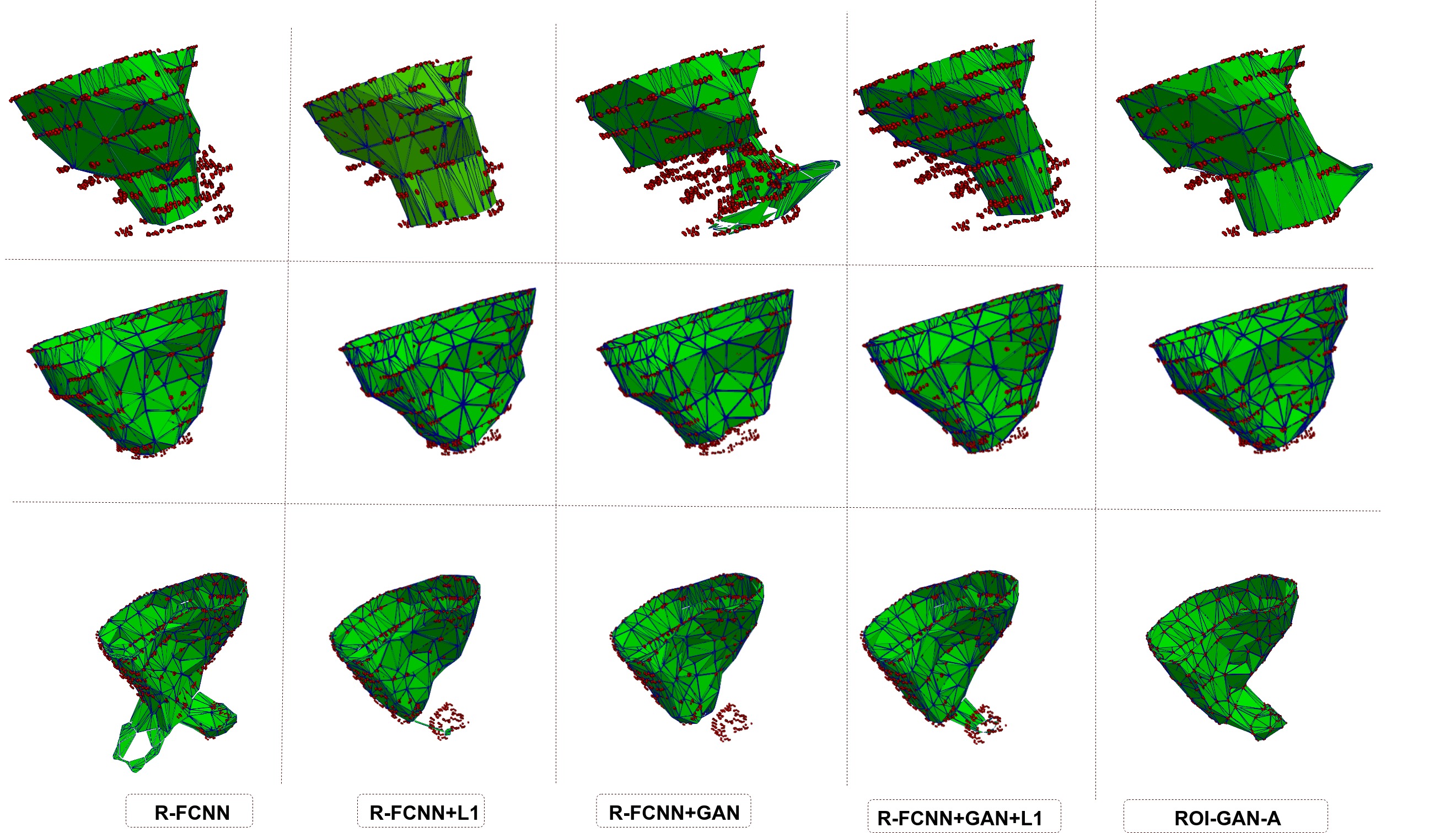}
   \caption{Examples of the reconstructed 3D anatomies of the RV, where the model prediction (green surface) is compared to the ground truth contours (red points).}
  \label{fig:3drendering}
\end{figure*}

The baseline for this study is the performance of the FCNN to fully automatically segment the RV. This section presents the gradual improvement from this baseline by applying the concepts of recurrence, L1 loss, GAN and finally the proposed ROI-GAN. Results will show how these concepts do not always complement each other. Illustrative examples of the segmentation performance are provided in Fig. \ref{fig:twinsuk} and Fig. \ref{fig:3drendering}. 

\subsection{Metrics}

Segmentation performance is evaluated for endocardial contours with two different metrics: Dice Index (DI) and Hausdorff Distance (HD). 
The DI is defined as twice intersection over the sum of cardinalities (i.e. a number of elements) of two binary masks $A$ and $B$,

$$
DI = \frac{2 | A \cdot B|}{|A| + |B|}.
$$ 

The HD is the greatest of all the distances from a point in one set to the closest point in the other set. It is defined as the $max(d_{a},d_{b})$, where $d_{a}$ is the distance from the automatic contour points to the closest point of the manual contour, and $d_{b}$ is the opposite. HD is measured in $mm$ in this work.

These metrics are provided for three anatomical regions of the heart because of the different challenges that they face. The middle of the right ventricle is the most stable and consistent part, but the top (base of the ventricle) suffers from quite a bit of variability in the shape of contours, and the bottom (apex of the ventricle) usually renders the worst performance due to the small size of the contour and poor contrast in the image. 

\subsection{The added value of a recurrent unit, GAN and L1 loss}

\begin{figure*}[ht]
\centering
\includegraphics[width=2.86in]{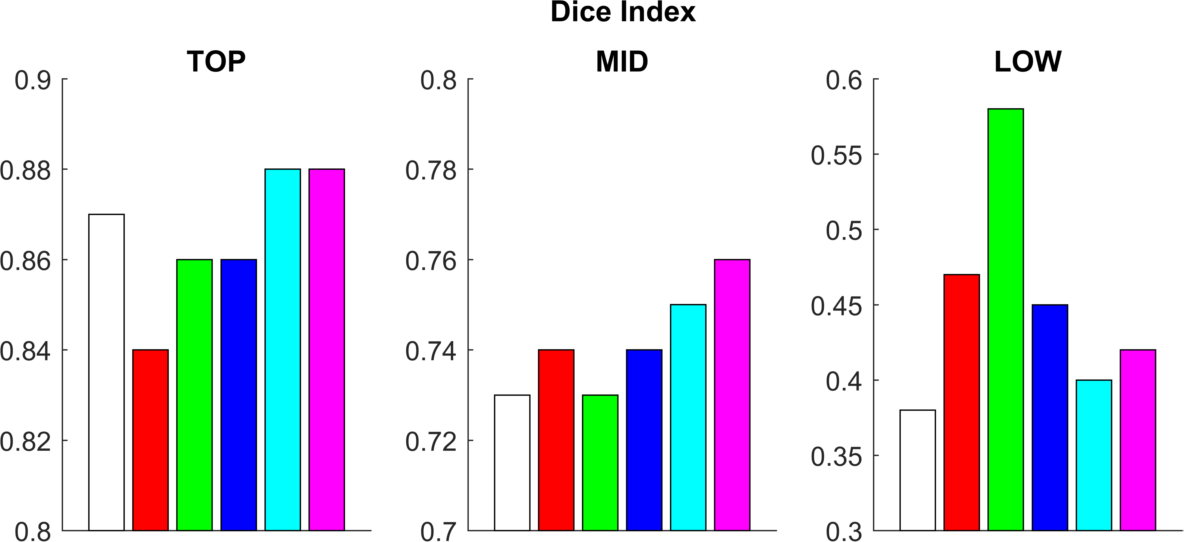}
\includegraphics[width=3.5in]{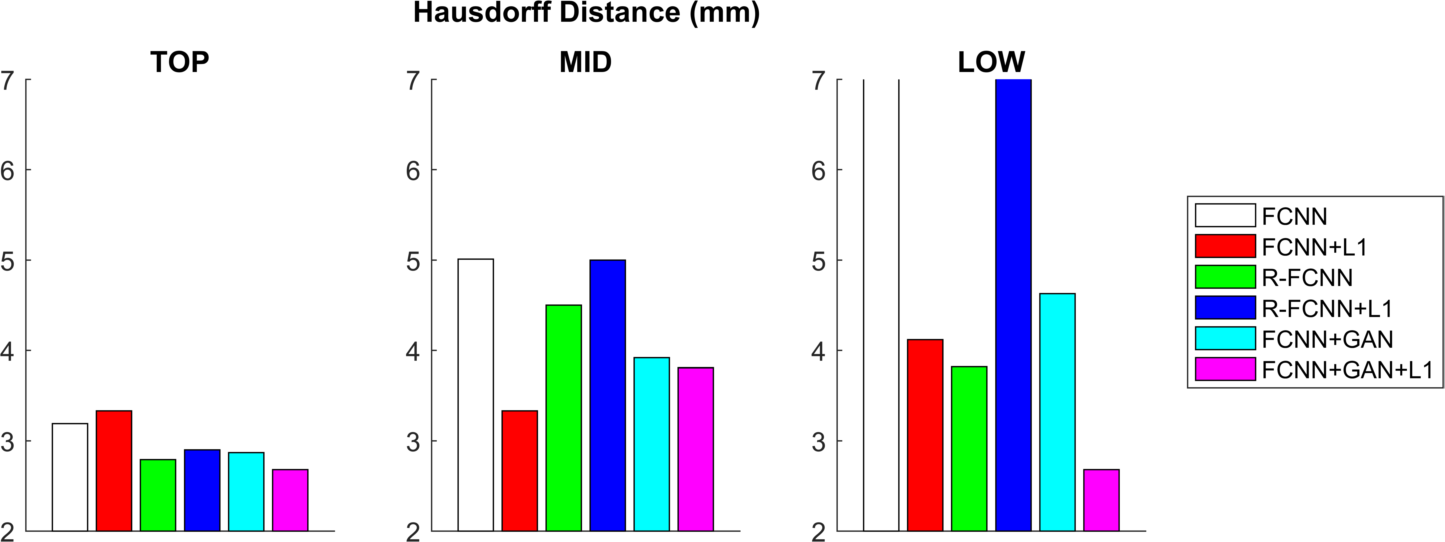}

   \caption{Evaluation of the added value of a recurrent unit (R-FCNN), the adversarial training (FCNN+GAN), and their combination with the L1 loss. Note that the HD bars in the LOW region for FCNN and R-FCNN+L1 reach larger values than the ones displayed in the plot.}
  \label{fig:section_a}
\end{figure*}

Fig. \ref{fig:section_a} shows how the R-FCNN introduces a significant improvement over the FCNN in the RV apical (low) region, with an increase of DI of a 52\% (from $0.38$ to $0.58$), and a reduction of the HD in a 72\% (from $13.60$ to $3.82$). The other two regions, top and mid, show a similar performance being only slightly worse at the DI of the top region. 

The addition of the L1 loss to the baseline FCNN also improves the performance at the apical region of the RV, both in DI and HD, with a small gain in both metrics at the mid-region, but with a drop of 0.03 in DI at the top region.

Finally, the GAN-FCNN improves the performance with respect to the FCNN in all regions and metrics, but with a small impact (DI jumps of 0.01-0.02, HD reduced in 0.5-1 mm) except for the large reduction of the HD in the apical region. 

The combination of L1 and R-FCNN leads to worse results than using any of these two concepts in isolation at the apical region but matches the best performance of the other two concepts in the other two regions. 

On the contrary, the combination of L1 and GAN leads to better results than using any of these two concepts in isolation, in all regions and using both metrics. The best performance this is the one provided by the FCNN+GAN+L1 and will be used as a baseline for the next experiments. 

Fig. \ref{fig:section_b} shows how the GAN and R-FCNN combination does not have any benefit, and that the performance of adding the L1 loss is even worse. 

 \begin{figure*}[ht]
\centering
 \includegraphics[width=7in]{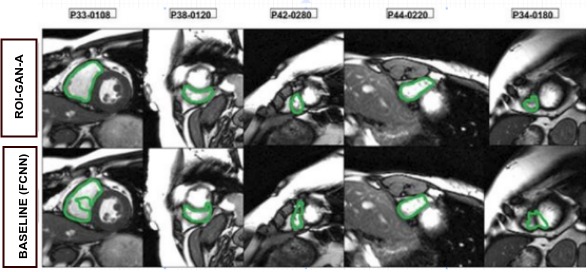}
   \caption{Examples of automatic segmentation results in images from the RV MICCAI dataset. These are instances where ROI-GAN-A showed superior segmentation results in comparison with the baseline FCNN.}
  \label{fig:rvpublic}
\end{figure*}       

\subsection{ROI-GAN with an R-FCNN provides the best performance}

\begin{figure*}[ht]
\centering
\includegraphics[width=2.86in]{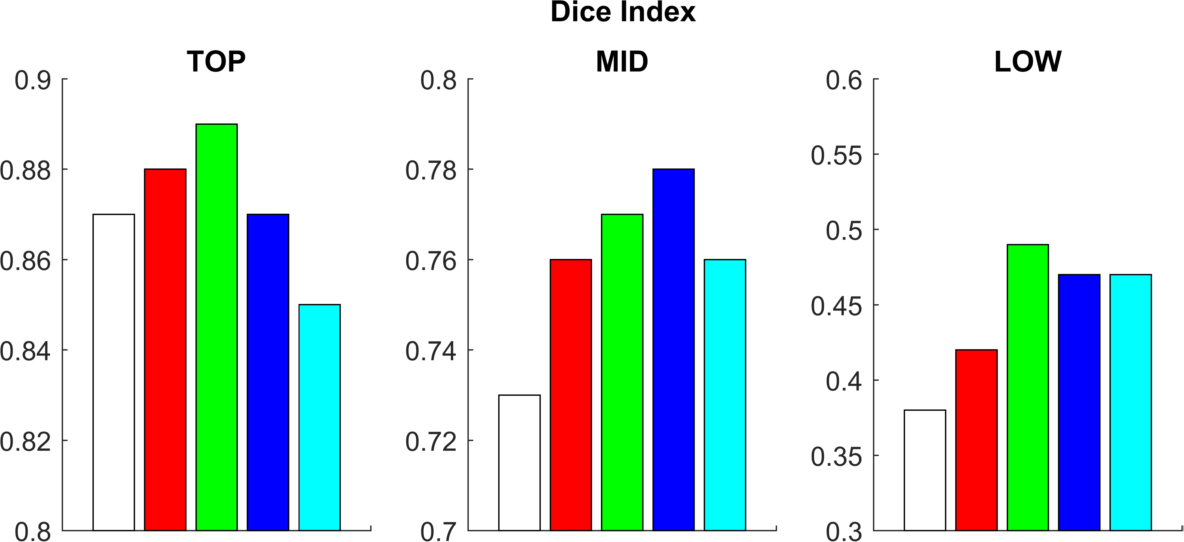}
\includegraphics[width=3.5in]{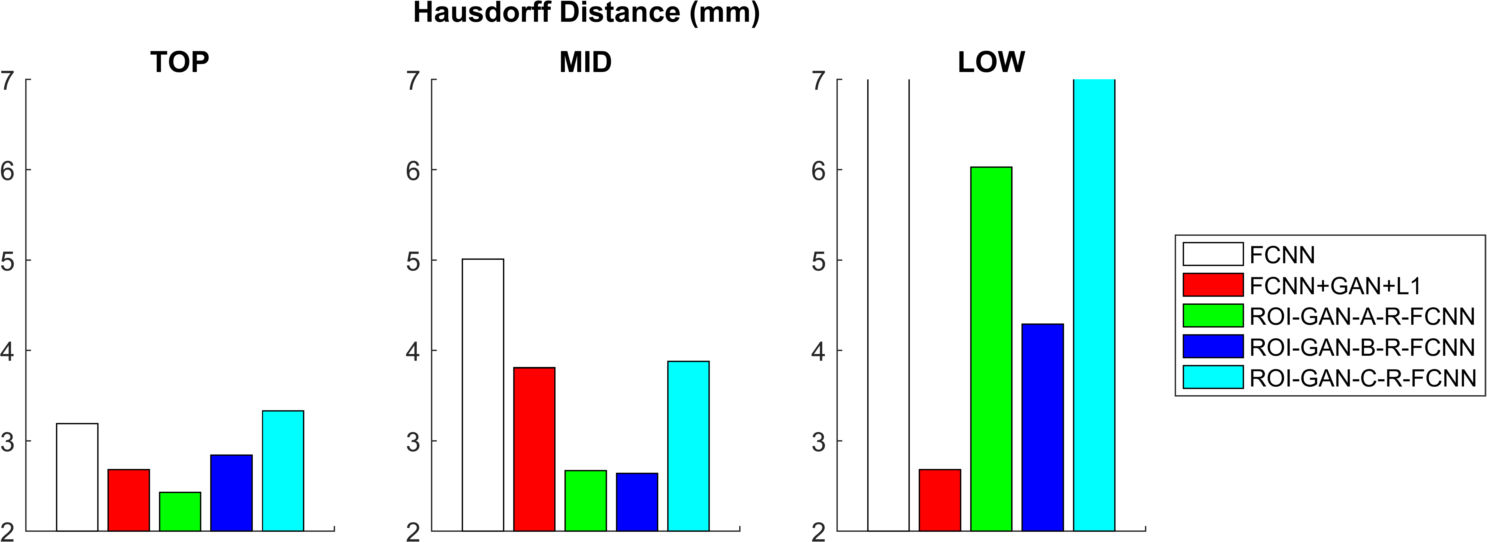}

   \caption{Benefit of the ROI-GAN over the baselines FCNN and FCNN+GAN+L1. Note how the gain from an FCNN to an FCNN+GAN+L1 is doubled with an ROI-GAN-A with an R-FCNN in all metrics but the HD of the low apical region.}
  \label{fig:section_c}
\end{figure*}

\begin{figure*}[ht]
\centering
\includegraphics[width=2.86in]{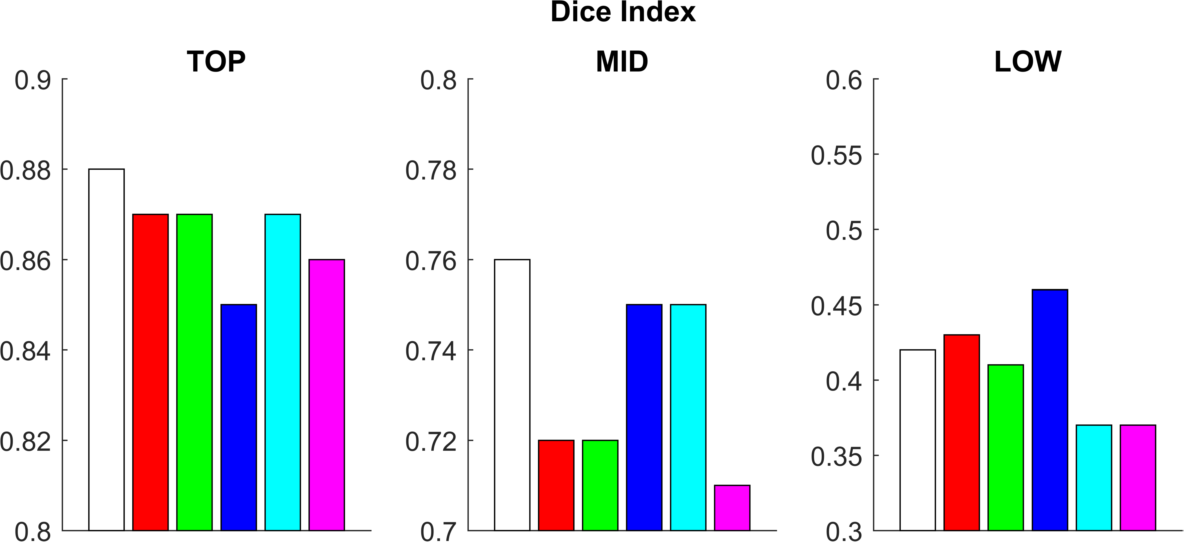}
\includegraphics[width=3.5in]{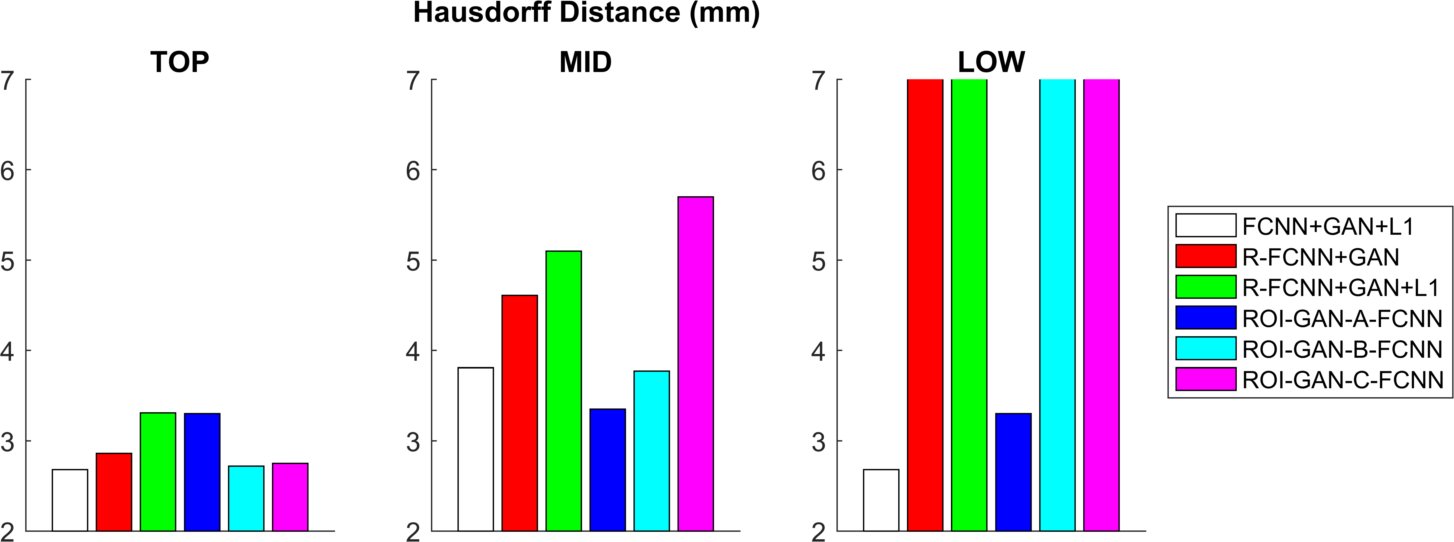}

   \caption{Strategies that do not improve the FCNN+GAN+L1 performance: the addition of the recurrent unit, or the ROI-GAN without a recurrent unit.}
  \label{fig:section_b}
\end{figure*}

The three versions of the ROI-GAN architecture are first evaluated with an FCNN as the generator, showing a drop in performance with respect to the FCNN+GAN+L1, see Fig. \ref{fig:section_b}. On the contrary, the ROI-GAN using R-FCNNs as generators improve the performance, see Fig. \ref{fig:section_c}, being the ROI-GAN-A the best in average in all regions.

In more detail, ROI-GAN-A delivers the best in all scores but in the DI of the mid-region, where ROI-GAN-B gets the best results, and the HD in the low apical region, where the FCNN+GAN+L1 delivers the best score.

\subsection{Generalization of results}

To confirm the results obtained in our in-house dataset, the performance of the baseline FCNN and of proposed ROI-GAN-A  architecture are tested in the RV MICCAI Challenge 2012. Illustrative examples are provided in Fig. \ref{fig:rvpublic}.

Table \ref{table:presu} shows how the ROI-GAN-A clearly improves over the  FCNN in DI and HD, with gains of $0.05$ and of $5.09$ mm respectively. This improvement in performance is of similar magnitude than the one observed in our in-house dataset, where an average of $0.05$ and $3.49$ mm was observed. 

The performance of the ROI-GAN-A, a fully automatic method, is close to the best methods of the literature, some of them semi-automatic (see Table \ref{table:presu}), which are able to further improve the DI in $0.03$ and reduce the HD in $0.75$mm). 

One last test is performed to evaluate the robustness in the extraction of clinical indexes such as volume or EF: a linear regression analysis between manual and automated endocardiac areas, for both ROI-GAN-A and  FCNN, reveal a $R$ correlation coefficient of $0.9642$ and $0.8899$ respectively.

\begin{table}[]
\centering
\begin{tabular}{|l|l|l|l|}
\hline
Methods               & FA/SA** & DM          & HD            \\ \hline
ROI-GAN-A & FA      & 0.80 (0.22) & 8.03(4.41)    \\ \hline
FCNN [Our baseline]  & FA      & 0.75(13.12) & 13.12(10.36)  \\ \hline
Avendi et al. \cite{MRAvendi}        & FA      & 0.82 (0.16) & 8.03 (4.41)   \\ \hline
Ringenberg et al 2014 \cite{Ringenberg} & FA      & 0.83 (0.18) & 8.73 (7.62)   \\ \hline
Zuluaga et al 2013  \cite{Zuluaga}   & FA      & 0.73 (0.27) & 12.50 (10.95) \\ \hline
Wang et al 2012 \cite{Wang}      & FA      & 0.61 (0.34) & 22.20 (21.74) \\ \hline
Ou et al 2012 \cite{Ou}        & FA      & 0.61 (0.29) & 15.08 (8.91)  \\ \hline
Maier et al 2012 \cite{Maier}     & SA      & 0.77 (0.24) & 9.79 (5.38)   \\ \hline
Nambakhsh et al 2013 \cite{Nambakhsh} & SA      & 0.56 (0.24) & 22.21 (9.69)  \\ \hline
Bai et al 2013 \cite{Bai}       & SA      & 0.76 (0.23) & 9.77 (5.59)   \\ \hline
Grosgeorge et al 2013 \cite{Grosgeorge} & SA   & 0.81 (0.16) & 7.28 (3.58)   \\ \hline
\end{tabular}
\newline
\caption{Segmentation performance results on the RV Test2Set MICCAI of public dataset. DI: Dice Index; HD: Hausdorff Distance (mm); FA: fully automatic; SA: semi automatic.}
\label{table:presu}
\end{table}

\begin{table*}[ht]
\centering
\label{my-label}
\begin{tabular}{ll|l|l|l|l|l|l|}
\cline{3-8}
                                                &                  & \textit{\textbf{TOP}} & \textit{\textbf{TOP}} & \textit{\textbf{MID}} & \textit{\textbf{MID}} & \textit{\textbf{LOW}} & \textit{\textbf{LOW}} \\ \hline
\multicolumn{1}{|l|}{\textbf{METHODS}}          & \textbf{FA/SA**} & \textbf{DM}           & \textbf{HD}           & \textbf{DM}           & \textbf{HD}           & \textbf{DM}           & \textbf{HD}           \\ \hline
\multicolumn{1}{|l|}{FCNN}                      & FA               & 0.87(0.20)            & 3.19(6.20)            & 0.73(0.28)            & 5.01(12.98)           & 0.38(0.37)            & 13.60(22.79)          \\ \hline
\multicolumn{1}{|l|}{\textbf{R-FCNN}}           & FA               & 0.86(0.21)            & 2.79(3.03)            & 0.73(0.28)            & 4.50(10.76)           & \textbf{0.58(0.30)}   & \textbf{3.82(10.83)}  \\ \hline
\multicolumn{1}{|l|}{FCNN+L1}                   & FA               & 0.84(0.25)            & 3.33(6.12)            & 0.74(0.26)            & 3.33(8.78)            & 0.47(0.36)            & 4.12(10.30)           \\ \hline
\multicolumn{1}{|l|}{R-FCNN+L1}                 & FA               & 0.86(0.24)            & 2.90(2.79)            & 0.74(0.27)            & 5.00(12.66)           & 0.45(0.36)            & 8.38(17.44)           \\ \hline
\multicolumn{1}{|l|}{FCNN+GAN}                  & FA               & 0.88(0.17)            & 2.87(4.65)            & 0.75(0.27)            & 3.92(10.03)           & 0.40(0.36)            & 4.63(9.06)            \\ \hline
\multicolumn{1}{|l|}{R-FCNN+GAN}                & FA               & 0.87(0.20)            & 2.86(5.57)            & 0.72(0.28)            & 4.61(12.33)           & 0.43(0.36)            & 9.17(18.50)           \\ \hline
\multicolumn{1}{|l|}{FCNN+GAN+L1}               & FA               & 0.88(0.18)            & 2.68(3.99)            & 0.76(0.25)            & 3.81(10.44)           & 0.42(0.38)            & 2.68(3.99)            \\ \hline
\multicolumn{1}{|l|}{R-FCNN+GAN+L1}             & FA               & 0.87(0.20)            & 3.31(7.63)            & 0.72(0.27)            & 5.10(13.55)           & 0.41(0.33)            & 14.46(25.01)          \\ \hline
\multicolumn{1}{|l|}{ROI-GAN-A-FCNN}            & FA               & 0.85(0.22)            & 3.30(6.61)            & 0.75(0.27)            & 3.35(8.64)            & 0.46(0.35)            & 3.30(6.61)            \\ \hline
\multicolumn{1}{|l|}{\textbf{ROI-GAN-A-R-FCNN}} & FA               & \textbf{0.89(0.18)}   & \textbf{2.43(2.21)}   & 0.77(0.22)            & 2.67(6.67)            & 0.49(0.33)            & 6.03(14.49)           \\ \hline
\multicolumn{1}{|l|}{ROI-GAN-B-FCNN}            & FA               & 0.87(0.21)            & 2.72(3.41)            & 0.75(0.27)            & 3.77(9.76)            & 0.37(0.37)            & 9.07(17.23)           \\ \hline
\multicolumn{1}{|l|}{\textbf{ROI-GAN-B-R-FCNN}} & FA               & 0.87(0.22)            & 2.84(4.60)            & \textbf{0.78(0.22)}   & \textbf{2.64(6.42)}   & 0.47(0.35)            & 4.29(9.89)            \\ \hline
\multicolumn{1}{|l|}{ROI-GAN-C-FCNN}            & FA               & 0.86(0.23)            & 2.75(2.75)            & 0.71(0.29)            & 5.70(14.49)           & 0.37(0.37)            & 11.56(20.51)          \\ \hline
\multicolumn{1}{|l|}{ROI-GAN-C-R-FCNN}          & FA               & 0.85(0.25)            & 3.33(7.24)            & 0.76(0.25)            & 3.88(10.93)           & 0.47(0.364)           & 7.29(16.56)           \\ \hline
\end{tabular}
\newline
\caption{Segmentation performance results on the Twins-UK dataset. DI: Dice Index; HD: Hausdorff Distance (mm); FA: fully automatic; SA: semi automatic.}
\label{table:twinstabresults}
\end{table*}
  
\section{Discussion}

The performance of the FCNN for the task of RV segmentation has been improved by the combination of the three existing concepts (R-FCNN, FCNN-GAN, L1 norm) through the ROI-GAN, a novel interpretation of the coGAN where two GANs are set to cooperate at two fields of view (general and local, or full resolution and ROI-focus). The best combination of the three existing concepts achieved an improvement of $0.05$ and $3.49$ mm in DI and HD respectively with respect to the baseline FCNN and this improvement was doubled with the ROI-GAN-A architecture. 

The combination of local and global features through an ROI-GAN has thus provided a benefit. The rationale sought was to enable the global FCNN to learn the useful features through the help of the local FCNN, coordinating their training with a generative adversarial game and sharing parameters (i.e. a CoGAN). CoGANs were proposed to avoid the need for paired datasets during the learning stage and reported excellent performance in the problem of Unsupervised Domain Adaptation \cite{Ming1}. The problem of image segmentation at two different fields of views in this work do share many more features than images from very different modalities (i.e. depth and color image, as in \cite{Ming1}), and we interpret this to be the main reason why sharing parameters was actually only beneficial at the generator, and not at the discriminator of the GANs (performance of ROI-GAN-A being superior to the ROI-GAN-C).

An interesting experimental finding was that the ROI-GAN-A architecture achieved the constructive coordination of the R-FCNN and GAN, which otherwise would result in a drop in segmentation performance (see Table \ref{table:twinstabresults}). The GANs have been proved to be strategical for enhancing the learning generalization (i.e. a better loss regularisation) \cite{DPathak, PhillipIsola}, and an R-FCNN models the spatial coherence as a set of connections to the previous slices (data input prepared with image slices from top to bottom of the heart) - these two concepts are not in apparent conflict, and the reason why these to concepts did not work together in a single FCNN, remains elusive to us. A possible explanation is the strong penalty that the GAN term imposes on MSE loss, that will be detrimental for a proper back-propagation through the C-GRU unit.

In this work, we illustrate that the problem of RV segmentation is, in fact, the combination of 3 problems, each of the 3 sections of the RV presents different challenges. While the top basal slices present a minor difficulty in localization but a great anatomical variability, the bottom apical slices present a great difficulty in localization but a much simpler shape (i.e RV collapses towards a circular structure), and mid slices will be an intermediate problem. As a consequence, there is not a single architecture being the optimal solution for all these 3 problems. In the top and middle slices, an ROI-GAN solution outperforms the rest (A or B configurations for top or mid slices), but the best solution to capture the apex of the RV is the R-FCNN. Further research is needed to design the architecture that adapts to the anatomical region present in the image. 

Redundancy across space, or time \cite{NSavioliLV},  is clearly a useful resource in the segmentation task. Exploiting the spatial redundancy is the rationale of the R-FCNN, and our results confirm the initial findings at the apex of the Left Ventricle (LV), where the main gain was observed compared to the rest of the anatomy \cite{RudraP}. The RV has a much greater anatomical variability, but this did not prevent a recurrent unit to better constrain the segmentation in the challenging apical slices. Nevertheless, the use of recurrent unit may not be an optimal solution, since is continuously limited by the problem of vanishing gradient that may decrease the overall performances \cite{KyungHyun}. Further research is needed to study alternatives such as 3D FCNNs, where 3D convolution capture all the spatial coherence/redundancy \cite{NSavioliAtrium}. 

The evaluation methodology followed in this work was designed to examine the concepts proposed while minimizing possible confounding factors regarding the learning rate. All FCNNs had the same number of convolution layers, up-convolutions, ReLU and BN units. Besides, the number of epochs and the learning rate were equal in the two datasets used, where at every batch size each network take as input a sequence of consecutive SA images (or the corresponding number of stack images in an R-FCNN). Nevertheless, we cannot claim an independence of all confounding factors (i.e. the characteristics of the datasets used).

Initial evidence of the generality of our findings was provided by testing the final proposed solution in the best public dataset available to our knowledge, where a distinct improvement in comparison with the baseline FCNN was found (i.e. 0.80 (0.22) vs 0.75 (13.12) in DI and HD respectively). This experiment also showed that another solution, based on classical and simple concepts such as thresholding, still achieves better results \cite{Ringenberg}, motivating the hypothesis that a combination of classical and deep learning approaches is an interesting direction of further research. 

Future works could also explore the use of 3D convolutions within the ROI-GAN, where the 3D generators (i.e global and local) should extract the spatial information in the better way without any R-FCNN vanishing gradient problem \cite{KyungHyun}, or the idea of multiple generators able to see different cropping scales of the input MRI sequences.

\end{document}